\documentclass[11pt]{article}

\usepackage[preprint]{acl}

\usepackage{times}
\usepackage{latexsym}
\usepackage[most]{tcolorbox}

\usepackage{tikz}
\usetikzlibrary{tikzmark,calc}

\usepackage{amsfonts}
\usepackage{amssymb}
\usepackage{placeins}
\usepackage{dblfloatfix}

\usepackage[T1]{fontenc}
\usepackage[utf8]{inputenc}
\usepackage{microtype}
\usepackage{inconsolata}
\usepackage{graphicx}
\usepackage{booktabs}
\usepackage{amsmath} 
\usepackage{longtable}
\usepackage{multirow}

\usepackage{array}
\newcolumntype{V}{!{\vrule width 0.6pt}}

\usepackage[table]{xcolor}
\usepackage{booktabs}
\usepackage{colortbl}

\definecolor{mydarkgreen}{RGB}{102,189,99}
\definecolor{mygreen}{RGB}{166,217,106}
\definecolor{mylightgreen}{RGB}{217,239,139}
\definecolor{myyellow}{RGB}{254,224,139}
\definecolor{myorange}{RGB}{253,174,97}
\definecolor{myred}{RGB}{244,109,67}
\definecolor{promptgray}{RGB}{95,95,95}
\definecolor{promptblue}{RGB}{34,132,155}
\definecolor{promptpurple}{RGB}{126,76,172}

\newcommand{\heatDG}[1]{\cellcolor{mydarkgreen}\textbf{#1}}
\newcommand{\heatG}[1]{\cellcolor{mygreen}#1}
\newcommand{\heatLG}[1]{\cellcolor{mylightgreen}#1}

\newcommand{\heatO}[1]{\cellcolor{myorange}#1}
\newcommand{\heatR}[1]{\cellcolor{myred}#1}

\newtcolorbox{promptbox}[1]{
  enhanced,
  breakable,
  colback=white,
  colframe=promptgray,
  colbacktitle=promptgray,
  coltitle=white,
  fonttitle=\bfseries,
  title={#1},
  boxrule=0.8pt,
  arc=2pt,
  left=5pt,
  right=5pt,
  top=5pt,
  bottom=5pt,
  toptitle=2pt,
  bottomtitle=2pt,
  before skip=0.5em,
  after skip=0.75em
}
\newcommand{\prompttok}[1]{\textcolor{promptblue}{\texttt{#1}}}
\newcommand{\promptvar}[1]{\textcolor{promptpurple}{\texttt{#1}}}

\title{
Does VLA Even Know the Basics? Measuring Commonsense and World Knowledge Retention in Vision–Language–Action Models
}

\author{\normalfont
Nikita Kachaev\textsuperscript{*,1} \quad
Andrey Moskalenko\textsuperscript{*,2,3,4,5} \quad
Matvey Skripkin\textsuperscript{2,6} \quad
Nikita Kurlaev\textsuperscript{7} \\
Daria Pugacheva\textsuperscript{7,8} \quad
Albina Burlova\textsuperscript{7,9} \quad
Mikhail Kolosov\textsuperscript{10} \quad
Denis Shepelev\textsuperscript{2,5} \\
Andrey Kuznetsov\textsuperscript{2} \quad
Elena Tutubalina\textsuperscript{7,11} \quad
Aleksandr I. Panov\textsuperscript{1,10} \\
Alexey K. Kovalev\textsuperscript{1,10} \quad
Vlad Shakhuro\textsuperscript{2,4,5} \\
\\[-0.2em]
{\footnotesize
\begin{tabular}{c}
\textsuperscript{1}CogAI Lab \quad
\textsuperscript{2}FusionBrain Lab \quad
\textsuperscript{3}IAI MSU \quad
\textsuperscript{4}Lomonosov MSU \quad
\textsuperscript{5}NUST MISIS \\
\textsuperscript{6}Applied AI Institute \quad
\textsuperscript{7}HSE University \quad
\textsuperscript{8}Generalizable AI Systems \quad
\textsuperscript{9}ISP RAS \\
\textsuperscript{10}MIRAI \quad
\textsuperscript{11}Domain-specific NLP Group \\
\textsuperscript{*}Equal contribution
\end{tabular}
}
}

\begin{document}
\maketitle
\begin{abstract}
Embodied Vision–Language–Action (VLA) models are typically obtained by fine-tuning powerful pretrained VLMs on robotics data, yet it is unclear how much commonsense and factual knowledge they retain after adaptation.
Failures on knowledge-sensitive tasks are ambiguous, conflating missing knowledge with poor generalization of low-level control.
We introduce Act2Answer, a lightweight protocol that adapts VLM knowledge benchmarks to VLA evaluation by requiring agents to answer through action.
Each question becomes a short tabletop episode where the agent performs a single object-placement action to select among candidate answers, yielding an action-grounded success rate with reduced control confounds.
We curate a test suite of such environments across diverse commonsense and world-knowledge categories and introduce layerwise intent probing to localize answer-relevant information across the VLM backbone and action head.
In a large-scale study of 7 VLA models and 9 VLM baselines, we systematically rank models across categories, finding that VLAs show solid performance on simple concepts while exhibiting larger gaps on richer semantic categories relative to their source VLMs, that VQA co-training is associated with better knowledge retention, and that answer-relevant signals peak in middle VLA layers but attenuate in upper layers.
\textsc{Act2Answer} is available at \href{https://tttonyalpha.github.io/act2answer/}{tttonyalpha.github.io/act2answer}.

\end{abstract}

\section{Introduction}

\begin{figure}[t!]
    \centering
    \includegraphics[width=1\linewidth]{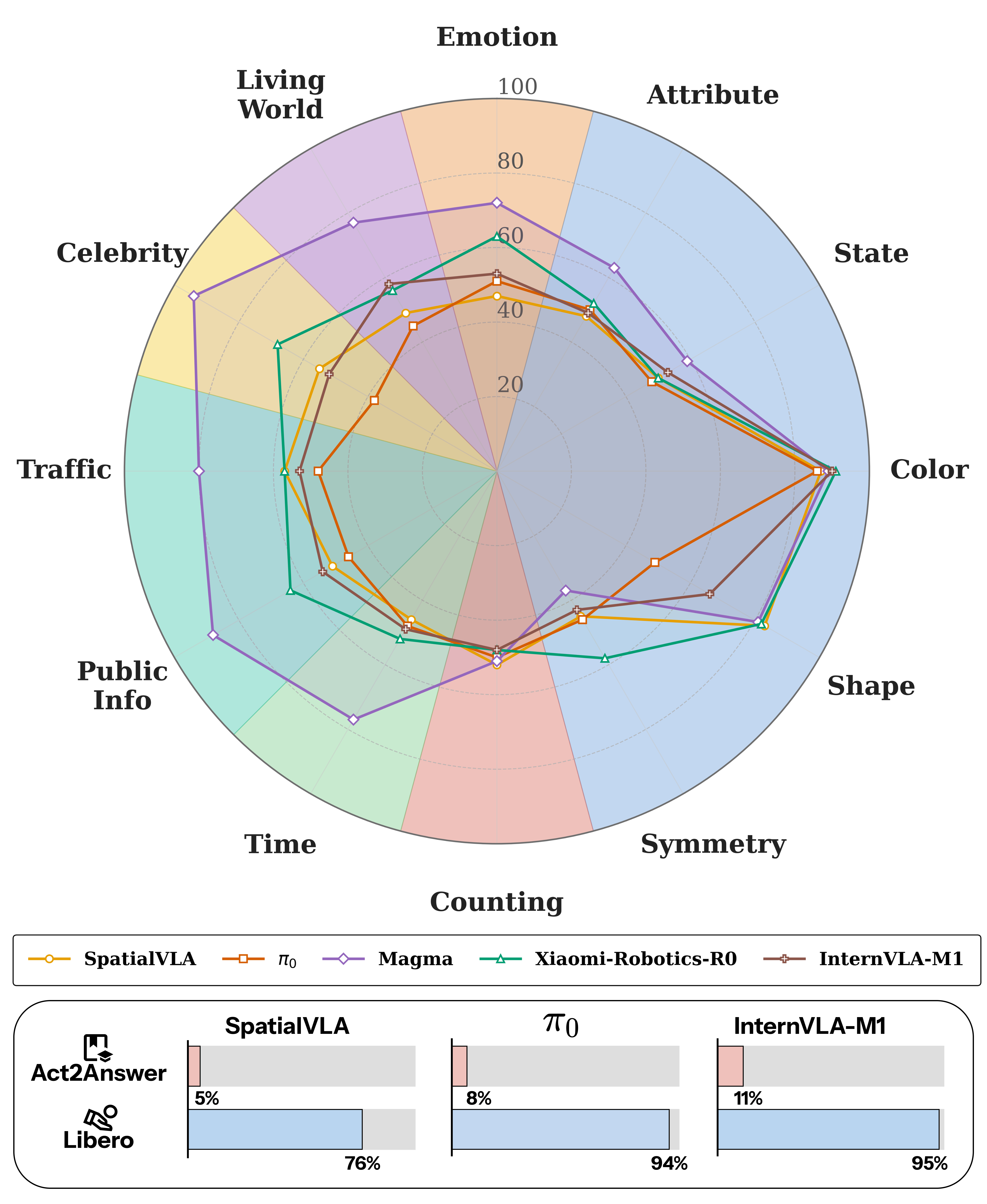}
    \caption{Knowledge evaluation results for seven state-of-the-art VLA models across diverse knowledge domains on the Act2Answer Task Suite. The bottom panel shows model performance averaged across all environments on both Act2Answer and LIBERO \cite{libero} (averaging details are provided in Appendix~\ref{appendix:score_averaging}).}
    \label{fig:intro}
\end{figure}

Embodied agents are increasingly studied as candidates for deployment in everyday environments, such as households~\cite{shukla2024maniskill, calvin, libero} and retail~\cite{soshin2025robobenchmart, liu2025fetchbot} settings. 
To operate effectively in such contexts, their actions must be grounded in a rich semantic understanding of the world - what objects are, how they are typically used, and which behaviors are appropriate in a given situation - tightly linking low-level motor control with commonsense, world-level reasoning.
Vision--Language--Action (VLA) models are widely proposed as the foundation for such agents and are often advertised as open-world-generalizable~\cite{pi0, openvla, magma, spatialvla, patratskiy2025spatial}, so we naturally expect them to handle previously unseen environments and objects with at least a basic level of appropriate generalization.

Yet, the rapidly expanding VLA literature has largely centered on manipulation-centric success.
Current benchmarks~\cite{libero, calvin, shukla2024maniskill, soshin2025robobenchmart, behavior1k, vlabench} ask whether agents can complete complex tasks under perturbations, domain shifts, or new layouts, but they rarely ask whether agents remain able to act on even basic commonsense distinctions about objects, scenes, and goals after robotics training.
Many VLA models are obtained by fine-tuning strong Vision-Language-Models (VLM) backbones on control tasks, and it is often implicitly assumed that the underlying world knowledge is preserved, even though we lack systematic ways to measure how much commonsense and factual knowledge is kept or catastrophically forgotten. In contrast, the VLM community has developed a rich ecosystem of benchmarks that explicitly probe world knowledge and commonsense. VLA evaluation, however, remains almost entirely task-success--centric: once a VLM backbone is fine-tuned into a policy that outputs actions, performance is usually reduced to success rates in manipulation or navigation domains, with little attention to whether the original knowledge is still present and accessible.
As a result, \textbf{we currently lack a principled way to test what a VLA model still knows after robotics fine-tuning.}

We introduce \textsc{Act2Answer}, an embodied evaluation protocol that adapts VLM knowledge benchmarks to VLA models by requiring action-based answer selection instead of text generation.
Each question becomes a short simulated episode with a simple selection action, reducing confounds from long-horizon planning and low-level control.
Prior action-based semantic evaluations~\citep{zitkovich2023rt,openvla} are limited in scope; in contrast, \textsc{Act2Answer} enables systematic benchmark adaptation across diverse commonsense and world-knowledge categories and supports layerwise analyses that localize answer-relevant information within VLA models. Specifically, our key contributions are as follows:
\begin{enumerate}
    \item We propose \textsc{Act2Answer}, an embodied evaluation benchmark suite that adapts VLM knowledge tasks into action-based simulated episodes, providing a controlled protocol for probing knowledge-sensitive behavior in VLAs through action rather than assessing VLAs text decoding on standard VLM QA benchmarks.
    
    \item By adapting existing VLM benchmarks, we curate a diverse embodied benchmark suite for systematically evaluating commonsense and world knowledge in VLA models.
    In total, we collect 1,720 unique binary questions across 12 categories, including attribute, state, color, symmetry, shape, emotion, celebrity, living world, counting, time, traffic, public info.
    
    \item We present a large-scale empirical study of 7 modern VLA systems and 9 strong VLM baselines, systematically ranking models across knowledge categories (Figure~\ref{fig:intro}).
    We find that current VLA models perform strongly on simple perceptual categories but exhibit substantially larger gaps on richer semantic categories relative to their source VLMs, and that VQA co-training is associated with stronger performance on knowledge-sensitive tasks.
    
    \item We introduce layerwise intent probing: linear classifiers trained on per-layer representations to predict the correct answer for each episode, to quantify how answer-relevant information is distributed across model depth.
    We show that relevant information can remain internally represented even when the model fails to translate it into the correct action.
\end{enumerate}

\section{Related Work}

\subsection{Evaluation of VLA Models}

Current VLA benchmarks primarily evaluate manipulation success, emphasizing control generalization across tasks, scenes, embodiments, and language variations rather than explicit knowledge assessment. Benchmarks such as LIBERO~\citep{libero}, CALVIN~\citep{calvin}, VLABench~\citep{vlabench}, RoboBenchMart~\citep{soshin2025robobenchmart}, and BEHAVIOR-1K~\citep{behavior1k} measure whether agents can complete language-conditioned tasks, while MIKASA-Robo~\citep{mikasa}, \citep{bringapplesofaimpact}, and VL-Think~\citep{blindvla} probe memory, robustness, and transfer. However, across these settings, evaluation remains largely grounded in manipulation success and mostly tests only shallow semantics or primitive concepts, leaving commonsense and world knowledge largely unmeasured.

Recent works have increasingly considered knowledge transfer from VLMs to VLA models. A common evaluation strategy, used in works such as \citep{cai2026xiaomi} and \citep{chen2025internvla}, is to test the VLM component on VQA-style benchmarks by decoding textual answers. This approach remains indirect: it measures whether the VLM component can still produce the correct textual answer, but not whether the VLA can use that knowledge when choosing an action. Consistent with this, \citet{zhang2026vlm4vla} show that strong VQA performance of VLM models does not necessarily translate into stronger VLA embodied behavior. In contrast, our Act2Answer protocol brings knowledge-sensitive evaluation into an embodied action setting, making failures more directly informative about missing task-relevant knowledge or the model’s inability to use it for correct action selection in context, rather than merely indicating whether the same information can still be verbalized by the underlying VLM: instead of answering in text, the agent must express its choice through action.

\subsection{Knowledge and Commonsense in VLM}

The VLM community has developed a broad set of benchmarks for evaluating multimodal understanding. Examples include GQA~\cite{hudson2019gqa} for compositional visual reasoning, TextVQA~\cite{singh2019towards} and DocVQA~\cite{mathew2021docvqa} for question answering that requires reading text in images, AI2D~\cite{kembhavi2016diagram} for diagram understanding, ScienceQA~\cite{lu2022learn} for multimodal science questions and MMMU~\cite{yue2024mmmu} for large-scale, multidisciplinary multimodal evaluation. These benchmarks have become standard tools for assessing the knowledge and reasoning capabilities of VLMs. It is therefore essential to use such established benchmarks when evaluating knowledge in VLA systems. Our Act2Answer protocol enables this by converting suitable items into an embodied binary-decision format, making it possible to test/probe knowledge-sensitive behavior in an action-based setting while retaining the benchmark grounding provided by existing VLM evaluations.

\begin{figure*}
    \centering
    \includegraphics[width=1\linewidth]{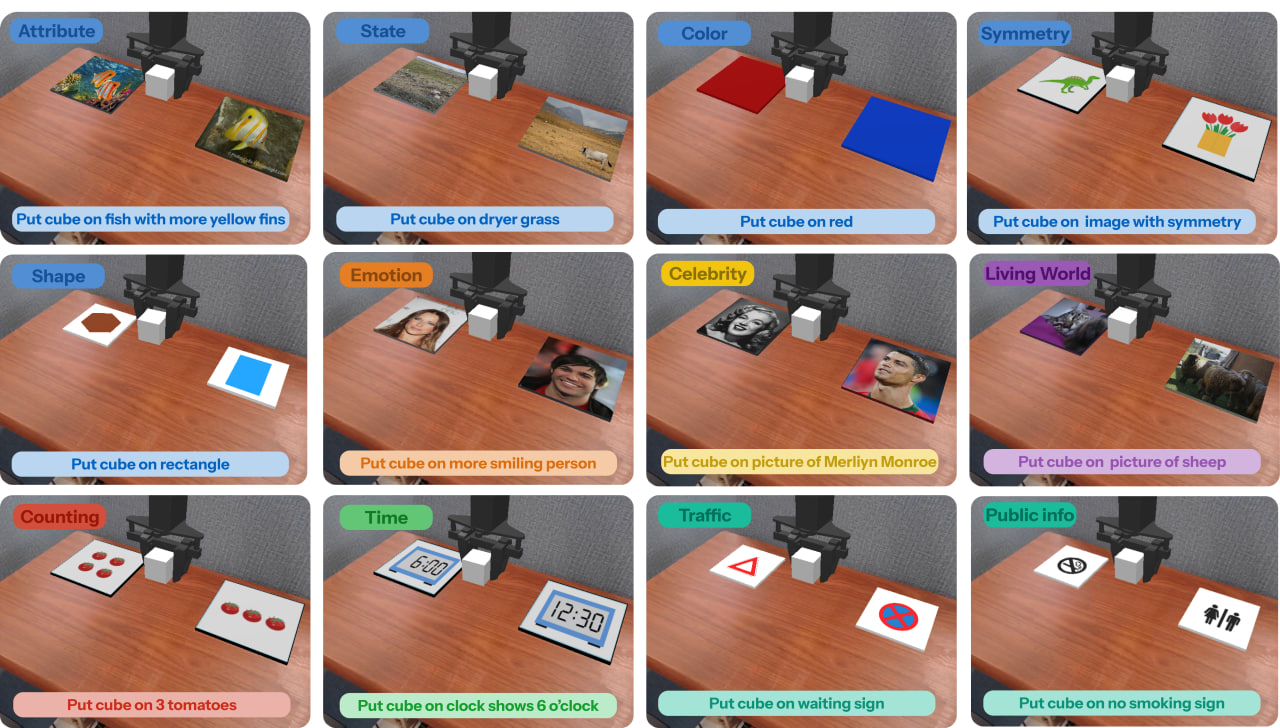}
    \caption{Act2Answer episodes examples for testing VLA models, built on top of VLM benchmark questions. In each episode, the embodied agent must interpret a natural-language instruction and control the robot arm to move the cube onto the correct answer plate.}
    \label{fig:act2answer_samples}
\end{figure*}

\section{Methodology}

\subsection{Decomposing Embodied Task Success}

Task success in embodied settings is not a single ability, but the outcome of several interacting factors. An agent must perceive the scene correctly, know which action is appropriate, execute that action reliably, and operate under the constraints of a particular environment. We use these four components: perception, knowledge, control, and environment - as a coarse conceptual decomposition of embodied task success. This decomposition matters because the same observed success rate may arise from very different underlying causes: strong knowledge but weak control, weak knowledge but strong motor routines, or even shortcut exploitation and favorable dynamics. As a result, end-to-end task success is often not diagnostic of what a VLA actually knows. Current benchmarks typically collapse these factors into a single outcome, making it difficult to tell whether failure reflects missing knowledge, perceptual error, motor difficulty, or environmental complexity. If the goal is to study knowledge in embodied agents, evaluation must therefore move beyond undifferentiated task success and more carefully isolate the contribution of knowledge from other sources of performance.

\subsection{Commonsense Knowledge}

What matters for embodied agents is not knowledge in the abstract, the question is what kinds of knowledge should actually be evaluated in embodied agents. Prior work such as Cosmos-Reason1~\citep{azzolini2025cosmos} has highlighted the importance of physically grounded reasoning over space, time, and causal constraints. However, real-world decision making often depends on a broader range of knowledge, including social roles, norms, quantities, biological constraints, and cultural context. 

In this work, we use the term \emph{commonsense knowledge} to refer to knowledge of this kind that can affect which action is appropriate in a given situation. Our goal is not to propose a universal taxonomy of embodied knowledge, but to introduce a practical set of knowledge categories that helps structure benchmark design and analysis. These categories are used to guide task selection, improve coverage across different types of commonsense knowledge, and support finer-grained error analysis. Because many embodied decisions are compositional, individual tasks may involve multiple factors at once. The proposed categories are therefore intended as a practical framework for organizing evaluation, rather than as a strict partition of embodied knowledge. Following this principle, we group the knowledge categories covered by our benchmark into seven broad domains:

\textbf{Physical} world knowledge covers properties of objects and materials, object states and state changes, object identity under occlusion or motion, spatial relations, affordances, intuitive mechanics, visibility and optics, thermodynamic processes, and physical causality or plausibility. These categories capture whether an agent can understand what objects are, what can be done with them, and what physical outcomes are possible. 
 
\textbf{Temporal} knowledge covers action semantics, event segmentation, temporal order, duration, delayed effects, goal–subgoal structure, and short-horizon prediction or planning. These categories capture whether an agent can interpret what is happening over time and choose an action that is appropriate not only for the current frame, but for the unfolding event. 

\textbf{Quantitative} knowledge covers counting, magnitude comparison, measurement, rates or proportions, and basic resource accounting such as available space, remaining time, or sufficient quantity. Many embodied decisions are not purely semantic: they depend on whether there is enough room, enough material, or the right relative amount to make an action succeed safely and correctly.

\textbf{Biological} knowledge covers distinctions between living and non-living entities, bodily vulnerability, injury risk, food safety, allergy or toxicity, animal and plant needs, and age- or development-dependent constraints. Such knowledge is often necessary for acting appropriately around humans, animals, and biologically meaningful objects.

\textbf{Social} knowledge covers agent roles, emotions, intentions, beliefs, relationships, communication, joint attention, cooperation, and conflict. In embodied settings, correct action often depends not only on physical state, but also on who the agents are, what they know, what they want, and how they are interacting.

\textbf{Normative} knowledge covers moral and conventional norms, safety rules, institutional rules, role-based obligations, contextual appropriateness, customs, symbols, and culture-dependent interpretation. These categories matter whenever multiple physically possible actions exist, but only some of them are socially acceptable, safe, or contextually appropriate.

\textbf{Cultural} knowledge covers shared cultural references and identities, such as well-known public figures, symbols, and conventions whose interpretation depends on broadly shared world knowledge rather than on the immediate physical or social scene. Such knowledge is what lets an agent recognize culturally salient entities and reason about them appropriately.

\begin{figure*}
    \centering
    \includegraphics[width=1\linewidth]{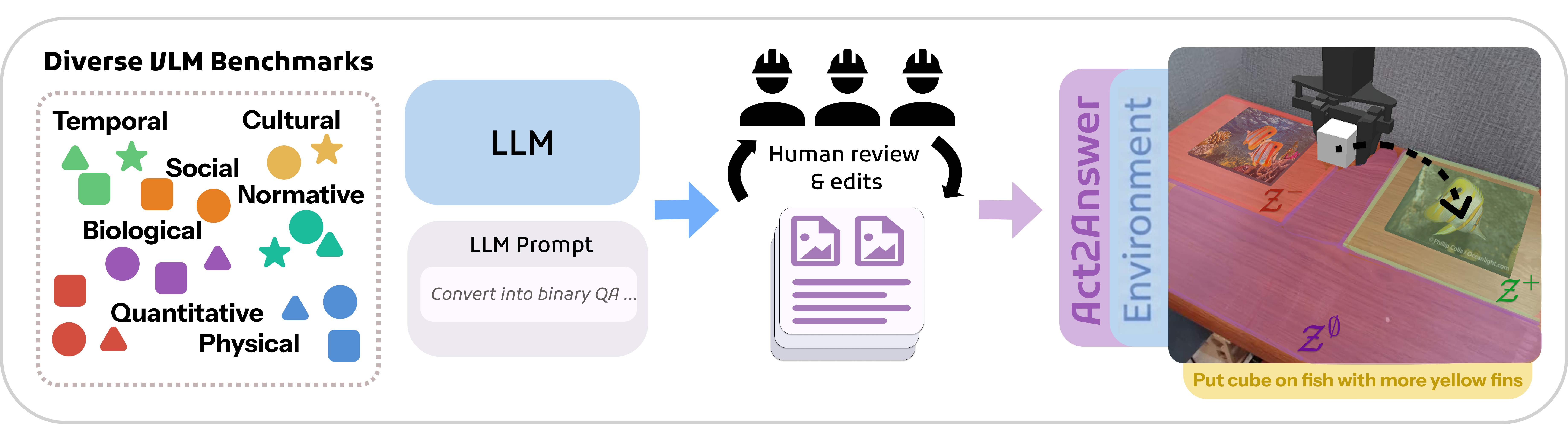}
    \caption{Overview of the data curation pipeline used to construct the Act2Answer task suite from VLM benchmarks, including selection, filtering and normalization, and conversion}    \label{fig:data_curation}
\end{figure*}

\section{Act2Answer: Embodied Evaluation of Knowledge}

The problem of knowledge evaluation in VLA models calls for a more direct way to assess \emph{action-relevant} world knowledge. Current benchmarks often entangle knowledge with perception, control, and environmental complexity, while text-based evaluation of the VLM part does not reveal whether that knowledge can still guide action. The key unresolved issue is whether a model can not only possess relevant knowledge, but also act on it. To address this, we introduce \textsc{Act2Answer}, a simple embodied evaluation protocol designed to evaluate whether an agent not only retains relevant world knowledge, but can also use it when selecting an action in a physically grounded setting. A useful parallel comes from cognitive science, where knowledge in nonverbal agents is often evaluated through knowledge-conditioned actions rather than verbal reports. Inspired by this paradigm, \textsc{Act2Answer} replaces textual answering with a minimal action that reveals the model’s choice.

\begin{table}[b]
\centering
\small
\begin{tabular}{ll}
\toprule
\textbf{VLM Benchmark} & \textbf{Domain} \\
\midrule
MLLM-CompBench & Emotion, Attribute, State \\
IconQA & Time, Shape, Symmetry, Counting \\
MMBench & Celebrity \\
OK-VQA & Living World \\
VL-Think & Public Info, Traffic, Color \\
\bottomrule
\end{tabular}
\caption{VLM benchmarks and knowledge domains used for adaptation to the Act2Answer protocol.}
\label{tab:vlm2vla_benchmarks}
\end{table}

\subsection{Evaluation Protocol}
The protocol is intentionally simple. Each episode is built from a VLM-style question together with one or more candidate images, such as a pair or a small grid, corresponding to possible answers (Figure~\ref{fig:act2answer_samples} shows representative episodes). These images are placed at known positions in the scene, and the textual instruction is provided to the agent in the standard way. The agent must indicate its answer through a minimal embodied response by moving its end-effector and placing a cube on the selected image. The prediction is scored as correct if the chosen image matches the correct answer. The protocol is designed to reduce the extent to which success is dominated by control difficulty or incidental environmental variation, and thereby make the outcome more directly informative about knowledge. \textsc{Act2Answer} is constructed so that success depends primarily on whether the agent can perceive, interpret the instruction, and use the required knowledge to select the correct answer, while minimizing the contribution of motor complexity and other low-level execution challenges. To this end, we deliberately reduce motor complexity: the required action is short-horizon, physically simple, and does not depend on difficult grasping or specialized manipulation skills. The resulting evaluation therefore does not fully isolate knowledge, but is intended to make the observed outcome more directly informative about whether the relevant knowledge remains available for action, rather than whether the agent can solve a difficult control problem.

\subsection{Tasks Description}

To evaluate VLA performance on the proposed knowledge-sensitive categories, we construct the \textsc{Act2Answer} task suite by adapting existing, community-established VLM benchmarks to an embodied setting rather than creating new question sets from scratch. This choice allows us to ground evaluation in widely used and already validated benchmark sources, while probing whether related knowledge-sensitive distinctions remain available for action after adaptation from VLM to VLA. For each broad knowledge category, we select a representative subcategory, \autoref{tab:vlm2vla_benchmarks} summarizes the correspondence between the source VLM benchmarks and the knowledge categories used in our evaluation. Each benchmark item is then converted into a standardized \textsc{Act2Answer} episode: the original question is reformulated as an instruction to the agent, and the candidate answers are instantiated as visual options placed in the scene. The agent must answer by performing the same minimal action defined in our protocol, namely selecting the correct option through object placement. 

\subsection{Data Curation}

Starting from a diverse pool of existing VLM benchmarks, we first select tasks that match the selected knowledge categories. Because current VLA models remain limited in long-context instruction following, we further filter the selected items by instruction length. We also apply image-level filtering: since many VLA models operate on relatively low visual resolution, human annotators remove examples in which the relevant objects are too small or visually ambiguous to be reliably perceived. After this selection stage, we unify the remaining tasks into a common embodied format. The original benchmark items may appear as open-ended questions or multiple-choice problems, so we convert them into binary decision tasks that are more suitable for action-based evaluation. To do so, we use an LLM to rewrite each example into a standardized two-option question while preserving its underlying knowledge requirement. Finally, these curated tasks are wrapped into an embodied environment built on the Simpler~\citep{simpler}, where each instance can be executed under the \textsc{Act2Answer} protocol. The result is a standardized task suite for evaluating whether VLA models retain and use Commonsense knowledge. In total, the suite covers 12 categories and contains 1,720 unique binary-choice items, corresponding to 3,440 evaluation episodes after including both original and swapped left/right configurations. \autoref{fig:data_curation} illustrates the data curation pipeline. For more details see \autoref{sec:appendix_benchmark_details}

\begin{table*}[t]
\centering
\footnotesize
\setlength{\tabcolsep}{4pt}
\renewcommand{\arraystretch}{0.95}

\caption{Results across knowledge-sensitive categories. VLAs (bottom) answer by embodied action selection under \textsc{Act2Answer}, VLM baselines (top) use the action-free text probe (RQ3).}
\label{tab:act2answer_emotion_attribute_state}

\resizebox{\textwidth}{!}{%
\begin{tabular}{lcccccccccccc}
\toprule
\textbf{Model}
& \multicolumn{1}{c}{\textbf{Social}}
& \multicolumn{5}{c}{\textbf{Physical}}
& \multicolumn{1}{c}{\textbf{Quantitative}}
& \multicolumn{1}{c}{\textbf{Temporal}}
& \multicolumn{2}{c}{\textbf{Normative}}
& \multicolumn{1}{c}{\textbf{Cultural}}
& \multicolumn{1}{c}{\textbf{Biological}} \\
\cmidrule(lr){2-2}
\cmidrule(lr){3-7}
\cmidrule(lr){8-8}
\cmidrule(lr){9-9}
\cmidrule(lr){10-11}
\cmidrule(lr){12-12}
\cmidrule(lr){13-13}
& \textbf{Emotion}
& \textbf{Attribute}
& \textbf{State}
& \textbf{Color}
& \textbf{Shape}
& \textbf{Symmetry}
& \textbf{Counting}
& \textbf{Time}
& \textbf{Public Info}
& \textbf{Traffic}
& \textbf{Celebrity}
& \textbf{Living World} \\
\midrule
InternVL3.5-8B
& \heatDG{95\%} & \heatG{68\%} & \heatG{64\%} & \heatDG{\textbf{100\%}} & \heatG{89\%} & \heatG{69\%} & \heatO{52\%} & \heatDG{99\%} & \heatG{85\%} & \heatG{75\%} & \heatDG{99\%} & \heatDG{91\%} \\

InternVL3.5-38B
& \heatDG{\textbf{99\%}} & \heatG{\textbf{73\%}} & \heatG{68\%} & \heatDG{\textbf{100\%}} & \heatDG{96\%} & \heatG{\textbf{83\%}} & \heatLG{59\%} & \heatDG{\textbf{100\%}} & \heatDG{\textbf{94\%}} & \heatG{81\%} & \heatDG{\textbf{100\%}} & \heatDG{96\%} \\

Ovis2.5-9B
& \heatG{89\%} & \heatG{69\%} & \heatG{\textbf{69\%}} & \heatDG{\textbf{100\%}} & \heatDG{\textbf{98\%}} & \heatG{\textbf{83\%}} & \heatLG{59\%} & \heatDG{99\%} & \heatG{88\%} & \heatG{85\%} & \heatDG{\textbf{100\%}} & \heatDG{\textbf{97\%}} \\

Qwen2.5-7B
& \heatG{89\%} & \heatG{64\%} & \heatG{68\%} & \heatDG{\textbf{100\%}} & \heatDG{90\%} & \heatG{78\%} & \heatG{62\%} & \heatDG{99\%} & \heatG{80\%} & \heatG{86\%} & \heatDG{\textbf{100\%}} & \heatDG{94\%} \\

Qwen2.5-32B
& \heatDG{\textbf{99\%}} & \heatG{69\%} & \heatG{\textbf{69\%}} & \heatDG{\textbf{100\%}} & \heatDG{93\%} & \heatG{\textbf{83\%}} & \heatG{61\%} & \heatDG{99\%} & \heatG{85\%} & \heatG{86\%} & \heatDG{\textbf{100\%}} & \heatDG{96\%} \\

Qwen3-8B
& \heatG{86\%} & \heatG{68\%} & \heatG{67\%} & \heatDG{\textbf{100\%}} & \heatDG{97\%} & \heatG{81\%} & \heatG{\textbf{65\%}} & \heatDG{98\%} & \heatG{83\%} & \heatDG{\textbf{93\%}} & \heatDG{\textbf{100\%}} & \heatDG{95\%} \\

Qwen3-32B
& \heatDG{92\%} & \heatG{67\%} & \heatG{66\%} & \heatDG{\textbf{100\%}} & \heatDG{\textbf{98\%}} & \heatG{\textbf{83\%}} & \heatLG{59\%} & \heatDG{\textbf{100\%}} & \heatG{87\%} & \heatDG{90\%} & \heatDG{\textbf{100\%}} & \heatG{86\%} \\

Prismatic-VLM-7B
& \heatG{82\%} & \heatLG{59\%} & \heatG{61\%} & \heatDG{96\%} & \heatG{85\%} & \heatG{67\%} & \heatO{52\%} & \heatDG{96\%} & \heatG{75\%} & \heatG{76\%} & \heatDG{99\%} & \heatG{82\%} \\

Paligemma-3B
& \heatO{53\%} & \heatO{50\%} & \heatO{52\%} & \heatO{47\%} & \heatO{48\%} & \heatO{49\%} & \heatO{49\%} & \heatO{51\%} & \heatO{48\%} & \heatO{49\%} & \heatO{48\%} & \heatO{49\%} \\
\midrule
OpenVLA
& \heatO{48\%} & \heatO{51\%} & \heatO{49\%} & \heatG{89\%} & \heatG{64\%} & \heatR{45\%} & \heatO{48\%} & \heatO{49\%} & \heatO{49\%} & \heatO{46\%} & \heatO{50\%} & \heatO{52\%} \\

OpenVLA (SFT)
& \heatR{41\%} & \heatR{45\%} & \heatR{44\%} & \heatG{82\%} & \heatO{53\%} & \heatR{38\%} & \heatO{46\%} & \heatR{37\%} & \heatO{47\%} & \heatO{46\%} & \heatR{42\%} & \heatR{45\%} \\

OpenVLA (RL)
& \heatO{46\%} & \heatO{50\%} & \heatO{47\%} & \heatG{88\%} & \heatG{61\%} & \heatR{44\%} & \heatO{50\%} & \heatO{48\%} & \heatO{46\%} & \heatO{48\%} & \heatO{47\%} & \heatO{52\%} \\

SpatialVLA
& \heatO{47\%} & \heatO{48\%} & \heatO{50\%} & \heatG{87\%} & \heatG{83\%} & \heatR{45\%} & \heatO{\textbf{52\%}} & \heatO{46\%} & \heatO{51\%} & \heatLG{57\%} & \heatLG{55\%} & \heatO{49\%} \\

$\pi_0$
& \heatO{51\%} & \heatO{50\%} & \heatO{48\%} & \heatG{86\%} & \heatO{49\%} & \heatO{\textbf{46\%}} & \heatO{50\%} & \heatO{48\%} & \heatO{46\%} & \heatO{48\%} & \heatR{38\%} & \heatR{45\%} \\

Magma
& \heatG{\textbf{72\%}} & \heatG{\textbf{63\%}} & \heatLG{\textbf{59\%}} & \heatG{89\%} & \heatG{\textbf{81\%}} & \heatR{37\%} & \heatO{51\%} & \heatG{\textbf{77\%}} & \heatG{\textbf{88\%}} & \heatG{\textbf{80\%}} & \heatDG{\textbf{94\%}} & \heatG{\textbf{77\%}} \\

Xiaomi-Robotics-R0
& \heatG{63\%}
& \heatO{52\%}
& \heatO{50\%}
& \heatDG{91\%}
& \heatG{82\%}
& \heatLG{58\%}
& \heatO{48\%}
& \heatO{52\%}
& \heatG{64\%}
& \heatLG{57\%}
& \heatG{68\%}
& \heatLG{56\%} \\

InternVLA-M1
& \heatO{53\%} & \heatO{49\%} & \heatO{53\%} & \heatDG{90\%} & \heatG{66\%} & \heatR{43\%} & \heatO{48\%} & \heatO{49\%} & \heatLG{54\%} & \heatO{53\%} & \heatO{52\%} & \heatLG{58\%} \\

\bottomrule
\end{tabular}%
}
\end{table*}

\subsection{Soft Success Rate}
\label{subsec:soft_success_rate}

To support consistent interpretation of VLA performance in \textsc{Act2Answer}, we define a simple interpretation methodology based on Soft Success Rate (SR). An episode is counted as successful if the agent places the cube on the correct answer tile within a tolerance region $\epsilon$ around it (\autoref{fig:data_curation}). To formally define this, let $p \in \mathbb{R}^2$ be the final 2D position of the cube at the end of an episode. We define a tolerance radius $\epsilon$ around the target image center $p^+$ and the incorrect option image center $p^-$ to partition the workspace $\mathcal{W}$ into target ($\mathcal{Z}^{+}$), incorrect ($\mathcal{Z}^{-}$), and
out-of-bounds (OOB) ($\mathcal{Z}^{\emptyset}$) regions as follows:
\begin{align}
\mathcal{Z}^+ &= \{p \in \mathcal{W} : \|p - p^+\| \le \epsilon\}, \\
\mathcal{Z}^- &= \{p \in \mathcal{W} : \|p - p^-\| \le \epsilon\}, \\
\mathcal{Z}^{\emptyset} &= \mathcal{W} \setminus (\mathcal{Z}^+ \cup \mathcal{Z}^-).
\end{align}
Then the Soft Success Rate (SR) over $N$ binary-choice tasks is the empirical estimate of the probability of landing in the target region:
\begin{equation}
\mathrm{SR} = \frac{1}{N}\sum_{i=1}^{N} \mathbb{I}\!\left(p^{(i)} \in \mathcal{Z}^+\right).
\end{equation}

Based on the probability mass distribution among these regions, performance can be interpreted in terms of three regimes defined around the random-guessing baseline of $0.5$. A score is treated as statistically distinguishable from chance only if it lies outside a chance-level interval of half-width $\Delta$, defined from binomial sampling fluctuations for the corresponding category. Details and category-specific values of $\Delta$ are given in \autoref{appendix:delta}.
\begin{enumerate}
    \item \textbf{Instruction or perceptual failure ($\text{SR} < 0.5 - \Delta$):} The model fails to ground the visual options or the instruction. This manifests either as a dominant probability of out-of-bounds placement that dwarfs the valid regions ($\mathbb{P}(p \in \mathcal{Z}^{\emptyset}) \gg \max\{\mathbb{P}(p \in \mathcal{Z}^+), \mathbb{P}(p \in \mathcal{Z}^-)\}$), or as a systematic semantic misunderstanding where the model consistently favors the incorrect option ($\mathbb{P}(p \in \mathcal{Z}^-) \gg \mathbb{P}(p \in \mathcal{Z}^+)$). In both cases, the probability mass in the target zone drops significantly below random chance.
    \item \textbf{No reliable usable knowledge ($|\text{SR} - 0.5| \le \Delta$):} The model correctly grounds the actionable regions ($\mathbb{P}(p \in \mathcal{Z}^{\emptyset}) \approx 0$) but lacks the specific knowledge required to select the correct option. This results in a near-uniform random guess or a severe positional bias between the two candidate regions, yielding $\mathbb{P}(p \in \mathcal{Z}^+) \approx \mathbb{P}(p \in \mathcal{Z}^-) \approx 0.5$ when evaluated across swapped spatial configurations.
    \item \textbf{Evidence of usable knowledge ($\text{SR} > 0.5 + \Delta$):} The model successfully grounds the actionable regions and correctly leverages its internal semantic knowledge to skew the action distribution toward the target zone, such that $\mathbb{P}(p \in \mathcal{Z}^+) > \mathbb{P}(p \in \mathcal{Z}^-)$.
\end{enumerate}
Finally, to reduce positional bias, we evaluate each example in both its original and swapped left/right versions and report the average score across the two. This makes the metric more robust to systematic side preferences.

\subsection{Linear Intent Probing}

Beyond task success, we measure whether answer-relevant information is linearly recoverable from a model's internal representations using \emph{layerwise intent probing}.
For each episode, we define a label $y \in \{0,1\}$ indicating the \emph{correct} answer option, i.e., the option region corresponding to the ground-truth answer rather than the model's own selection.
For a given VLA model, we extract hidden states from all tokens of every transformer layer, including both the VLM and Action Expert parts. Since the Action Expert can be iterative (for example, 10 steps of flow-matching), we extract activations from every iteration. For each VLA, we then train a linear probe independently on the activations of each layer to predict $y$ and report probe accuracy as a function of layer number. For Action Expert layers, we perform this procedure independently for all iterations. Rather than focusing on the absolute probe accuracies of the VLM and Action Expert parts in isolation, we focus on their relative relationship, which serves as an implicit indicator of how much answer-relevant information becomes attenuated along the path to action selection.

Formally, let $s_n^{\mathrm{bb}}$ denote probe accuracy at backbone layer $n$, and let $s_n^{\mathrm{exp}}$ denote probe accuracy at Action Expert layer $n$. Since our main interest is in above-chance recoverable signal, we summarize this relationship using chance-normalized quantities. First, we define the \textbf{Chance-Normalized Retention} as
\[
\mathrm{Retention}
=
\frac{\max_n (s_n^{\mathrm{exp}}-c)}{\max_n (s_n^{\mathrm{bb}}-c)+\varepsilon},
\]
where $c$ is the chance-level accuracy and $\varepsilon$ is a small constant for numerical stability. This metric compares the strongest above-chance probing signal in the Action Expert to the strongest above-chance probing signal in the backbone.

\subsection{Evaluation and Results}

Our goal is not to establish a new leaderboard, but to obtain an initial picture of current VLA performance on knowledge-sensitive categories under a controlled Act2Answer protocol. We evaluate several popular VLAs, including $\pi_0$~\citep{pi0}, OpenVLA~\citep{openvla}, Magma~\citep{magma}, Xiaomi-Robotics-R0~\citep{xiaomirobotics}, InternVLA-M1 \citep{chen2025internvla}, SmolVLA \citep{smolvla} and SpatialVLA~\citep{spatialvla}, and compare them to strong VLM baselines such as Qwen2.5-VL, Ovis, PaliGemma, and InternVL (\autoref{tab:act2answer_emotion_attribute_state}). Unless noted otherwise, we evaluate all VLA models using their original released checkpoints, without additional task-specific fine-tuning, the only exception is a separate set of supplementary ablations for OpenVLA. We then analyze the results to answer a set of research questions about what kinds of knowledge current VLAs preserve, lose, and remain able to use in action.

\paragraph{RQ1: How well do current VLA models handle simple primitives?}
We begin by testing whether VLA models can solve tasks built around basic perceptual concepts such as \textsc{Color} and \textsc{Shape}. These categories serve as a useful lower bound: if models fail here, it would suggest a broad inability to use even simple visual distinctions in action. In practice (\autoref{tab:act2answer_emotion_attribute_state}), nearly all evaluated models perform strongly on such tasks, with high success rates across primitive categories, although $\pi_0$ is a notable exception on \textsc{Shape}, where its success rate is close to chance. This indicates that simple basic physical and perceptual knowledge remains behaviorally accessible in current VLA systems.

\paragraph{RQ2: Can VLA models handle more complex semantic concepts?}
We next turn to categories that place greater demands on abstract semantic interpretation than primitive visual matching. Across the full set of non-primitive categories, current VLAs mostly remain at or near the random threshold, suggesting that once correct behavior depends on richer semantic, quantitative, temporal, normative, cultural, or biological distinctions, performance becomes highly unstable (\autoref{tab:act2answer_emotion_attribute_state}). On \textsc{Emotion}, \textsc{Attribute}, and \textsc{State}, \textsc{Time}, nearly all models are near chance, with Magma as the only clear exception. Particularly strikingly, no evaluated VLA reaches above-random performance on \textsc{Symmetry} or \textsc{Counting}, suggesting that these categories remain uniformly challenging for all tested VLA models. Normative, cultural, and biological categories show the same overall pattern. Most models remain close to chance on \textsc{Public Info}, \textsc{Traffic}, \textsc{Celebrity}, and \textsc{Living World}, whereas Magma again stands out with a large margin above threshold on all of them. Outside of Magma, gains are sparse and category-specific, such as SpatialVLA on \textsc{Traffic} (57\%) and \textsc{Celebrity} (55\%), or InternVLA-M1 on \textsc{Living World} (58\%). Overall, these results suggest that current VLAs struggle once correct action depends on more than shallow perceptual cues, and that richer semantic information is often not reliably available for action selection in the evaluated models.

\paragraph{RQ3: How large is the VLM-VLA gap?}
To obtain an upper-bound estimate of how much performance on knowledge-sensitive tasks may remain available after the transition from VLM to VLA, we compare each VLA to its original VLM checkpoint in an action-free probing setup. Given the first frame, we ask the VLM: \textit{``Do you see the <board\_name>? Answer yes or no. If yes, specify where: left, center, or right.''} A prediction is counted as correct only if both the board identity and its position match the ground truth, yielding a rough estimate of semantic grounding that can be related to Act2Answer performance. The results (\autoref{tab:act2answer_emotion_attribute_state}) reveal a substantial gap: across most domains, the original VLMs exceed their VLA counterparts by roughly 20-40 points. This provides evidence consistent with a marked drop in performance on knowledge-sensitive tasks after adaptation from vision-language pretraining to embodied policy learning.

\begin{figure}
    \centering
    \includegraphics[width=1\linewidth]{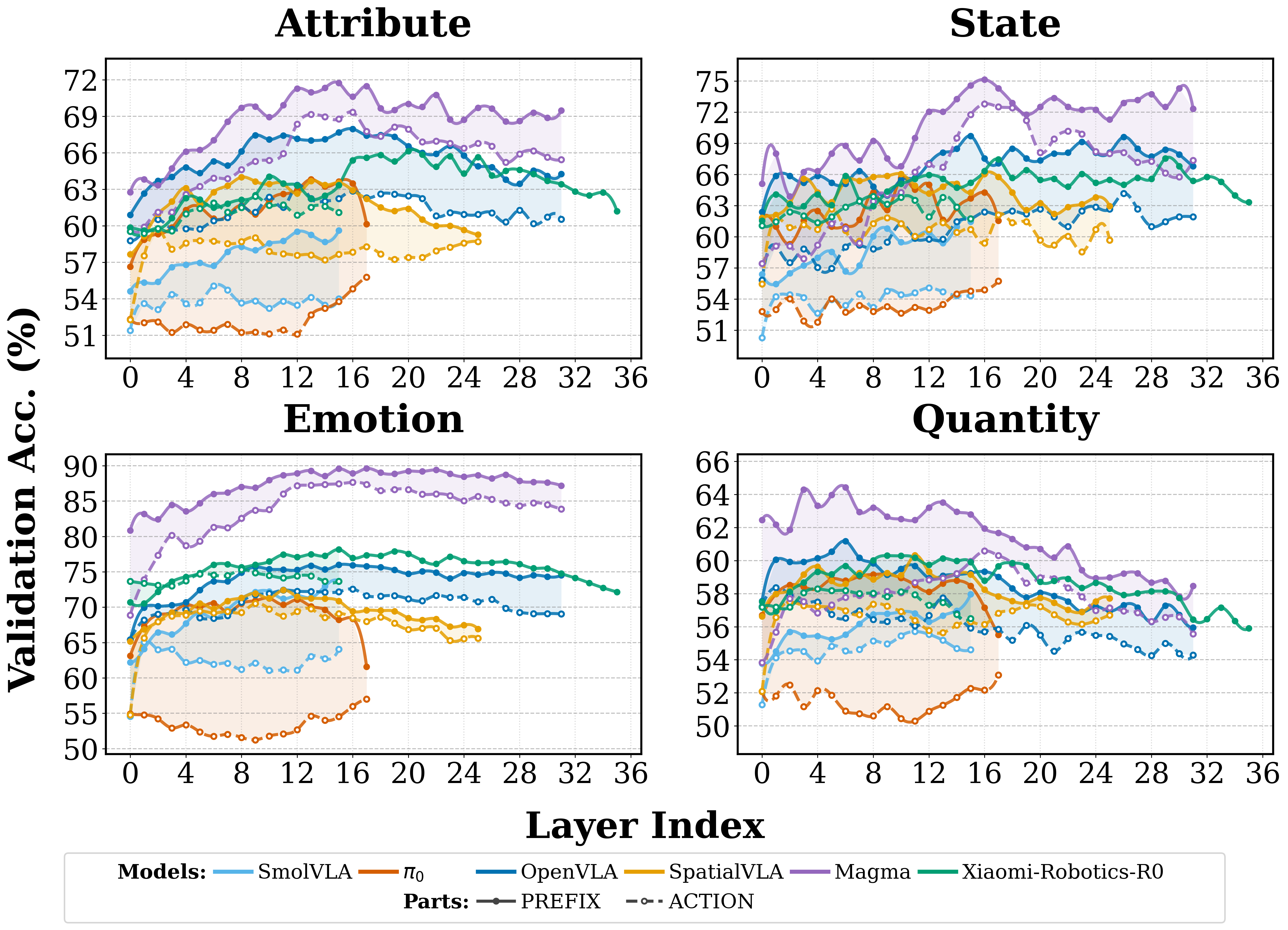}
    \caption{Probing results for internal representations of VLA models on four tasks from the Act2Answer task suite. In the legend, \texttt{Prefix} labels indicate representations from the VLM component, whereas \texttt{Action} labels indicate representations from the Action component.}
    \label{fig:probing}
\end{figure}

\paragraph{RQ4: Where does the knowledge go?}
A natural follow-up is whether answer-relevant information is truly erased or whether it remains somewhere in the hidden representations but is no longer accessible for action.
To study this, we perform linear probing over all layers of VLA models on categories such as \textsc{Attribute}, \textsc{State}, \textsc{Emotion}, and \textsc{Counting}.
The results in \autoref{fig:probing} reveal a consistent pattern: intermediate layers of the VLM backbone are often above chance, suggesting that task-relevant information is still present, but performance declines toward the final layers used for action prediction, often approaching random guessing.
This suggests a bottleneck between semantic representation and action generation: the model may retain answer-relevant information in intermediate representations, but fail to reliably translate it into the correct action.

\begin{table}[t]
\centering
\footnotesize
\setlength{\tabcolsep}{4pt}
\renewcommand{\arraystretch}{0.95}

\caption{Averaged probing-based retention metrics by model. VLM and Action report the maximum probing accuracies over backbone and Action Part layers, respectively.}
\label{tab:cnr_car}
\resizebox{\columnwidth}{!}{%
\begin{tabular}{lccc}
\toprule
\textbf{Model} & \multicolumn{2}{c}{\textbf{Probing Accuracy}} & \\
\cmidrule(lr){2-3}
& \textbf{Prefix} & \textbf{Action} & \textbf{Retention} \\
\midrule
Magma         & \textbf{75.23} & \textbf{72.60} & \textbf{0.8702} \\
Xiaomi-Robotics-R0        & 68.04 & 64.98 & 0.8159 \\
SpatialVLA & 65.70 & 62.60 & 0.7808 \\
OpenVLA    & 68.71 & 64.61 & 0.7697 \\
SmolVLA       & 63.18 & 57.73 & 0.5809 \\
$\pi_0$       & 64.99 & 55.40 & 0.3620 \\
\bottomrule
\end{tabular}%
}
\end{table}

\paragraph{RQ5: Does vision-language supervision improve knowledge-sensitive performance?}
We also ask whether continued vision-language supervision during VLA training is associated with stronger performance on knowledge-sensitive categories. To examine this, we explicitly include two groups of VLA baselines: models trained with joint vision-language and robotics supervision, such as Magma, Xiaomi-Robotics-R0, and InternVLA-M1, and models trained primarily or almost exclusively on robotics data, such as OpenVLA \citep{openvla}, SpatialVLA \citep{spatialvla}, and $\pi_0$ \citep{pi0}. The overall trend is positive: models in the first group perform better on average across most categories, especially on higher-level semantic, temporal, normative, cultural, and biological tasks, than models in the second group. While the gains are not uniform in every domain, this pattern suggests that continued vision-language supervision is associated with stronger action-grounded performance on knowledge-sensitive tasks and may help maintain task-relevant semantic information during embodied training.

\paragraph{RQ6: How does downstream fine-tuning affect knowledge-sensitive performance?}
Finally, we study whether additional downstream fine-tuning improves or harms performance on knowledge-sensitive categories. Using OpenVLA as a case study, we fine-tune the model with both SFT and RL on a small pick-and-place dataset that is not directly drawn from our benchmark, but does include visual and semantic perturbations. The results do not show consistent improvements. On the contrary, some categories, including \textsc{State} and \textsc{Color}, exhibit noticeable drops after SFT fine-tuning. This suggests that standard downstream adaptation may further bias the model toward task-specific action optimization, sometimes at the expense of more general knowledge-sensitive performance.

\section{Conclusion}

We introduced \textsc{Act2Answer}, a simple protocol for evaluating knowledge-sensitive behavior in VLA models.
In our setup, a VLA operates in a table-top environment with several candidate images and a natural-language prompt, and must answer by performing a minimal embodied action, such as placing a dummy object onto the image it believes to be correct.
This design keeps interaction embodied, but makes the choice itself as close as possible to the multiple-choice formats used in VLM benchmarks, so that failures are more directly informative about knowledge-sensitive behavior than about low-level motor difficulty.

Our investigation reveals a consistent gap between strong performance on simple perceptual categories and substantially weaker performance on richer semantic categories in current VLA systems.
The transition from VLM to VLA tends to preserve low-level visual discrimination (e.g., color, shape, coarse object identity), yet performance drops markedly on higher-level categories.
This pattern suggests that current training pipelines often preserve the ability to act on shallow perceptual cues while weakening performance on tasks that require richer task-relevant semantic distinctions.
A robot that can reliably grasp a cup but cannot distinguish a ``dirty'' cup from a ``clean'' one, or a ``sad'' human from a ``neutral'' one, is fundamentally limited in its usefulness as an assistant in everyday environments.
These results indicate that simply fine-tuning VLMs on action data is insufficient: the next generation of embodied agents will require architectures and training objectives that better maintain and align the backbone’s action-relevant semantic understanding with its learned motor policies, rather than allowing stronger control adaptation to come with weaker performance on broader knowledge-sensitive categories.

\bibliography{bibtex}
\clearpage

\appendix

\label{sec:appendix}

\section{Evaluation and Setup Ablations}
\label{sec:appendix_eval_setup_ablations}

This appendix reports ablations that test whether the main \textsc{Act2Answer} conclusions are sensitive to evaluation choices rather than knowledge-sensitive action selection.
We consider two ablations for the VLM comparison, image resolution and prompt formulation, and three ablations for the embodied evaluation setup, texture rendering, answer-tile size, and lighting intensity. These experiments help verify that the central findings are not driven by a single prompt, image preprocessing choice, or simulator rendering condition.

\subsection{Effect of Image Resolution}

The VLM--VLA comparison in the main paper is intended as an action-free estimate of how much task-relevant information remains accessible to the VLM backbone.
To check whether this comparison is dominated by image preprocessing, we evaluate VLM baselines under two image resolutions: $224 \times 224$ and $560 \times 480$.
As shown in Table~\ref{tab:appendix_vlm_resolution_prompt_ablation}, changing image resolution slightly affects some categories and models, especially fine-grained categories such as \textsc{Attribute}, \textsc{State}, and \textsc{Symmetry}.
However, the qualitative interpretation remains unchanged: strong VLM baselines continue to perform well on many categories where most VLA policies remain near chance, suggesting that the VLM--VLA gap is not solely an artifact of the chosen image resolution.

\begin{table*}[t]
\centering
\footnotesize
\setlength{\tabcolsep}{4pt}
\renewcommand{\arraystretch}{0.95}

\caption{VLM ablations over image resolution and prompt formulation. Prompt p1 is the perceptual QA-style prompt, and p2 is the VLA-style action prompt. These ablations test whether the VLM--VLA comparison is sensitive to image preprocessing or prompt wording.}
\label{tab:appendix_vlm_resolution_prompt_ablation}

\resizebox{\textwidth}{!}{%
\begin{tabular}{lccccccccccccccc}
\toprule
\textbf{Model} & \textbf{Resolution} & \textbf{Prompt}
& \multicolumn{1}{c}{\textbf{Social}}
& \multicolumn{5}{c}{\textbf{Physical}}
& \multicolumn{1}{c}{\textbf{Quantitative}}
& \multicolumn{1}{c}{\textbf{Temporal}}
& \multicolumn{2}{c}{\textbf{Normative}}
& \multicolumn{1}{c}{\textbf{Cultural}}
& \multicolumn{1}{c}{\textbf{Biological}} \\
\cmidrule(lr){4-4}
\cmidrule(lr){5-9}
\cmidrule(lr){10-10}
\cmidrule(lr){11-11}
\cmidrule(lr){12-13}
\cmidrule(lr){14-14}
\cmidrule(lr){15-15}
& & 
& \textbf{Emotion}
& \textbf{Attribute}
& \textbf{State}
& \textbf{Color}
& \textbf{Shape}
& \textbf{Symmetry}
& \textbf{Counting}
& \textbf{Time}
& \textbf{Public Info}
& \textbf{Traffic}
& \textbf{Celebrity}
& \textbf{Living World} \\
\midrule
InternVL3.5-8B
& \multirow{4}{*}{224$\times$224}
& \multirow{4}{*}{p1}
& \heatDG{95\%} & \heatG{68\%} & \heatG{64\%} & \heatDG{\textbf{100\%}} & \heatG{89\%} & \heatG{69\%} & \heatO{52\%} & \heatDG{99\%} & \heatG{85\%} & \heatG{75\%} & \heatDG{99\%} & \heatDG{91\%} \\

Ovis2.5-9B
& & 
& \heatG{89\%} & \heatG{69\%} & \heatG{\textbf{69\%}} & \heatDG{\textbf{100\%}} & \heatDG{\textbf{98\%}} & \heatG{\textbf{83\%}} & \heatLG{59\%} & \heatDG{99\%} & \heatG{88\%} & \heatG{85\%} & \heatDG{\textbf{100\%}} & \heatDG{\textbf{97\%}} \\

Qwen2.5-7B
& & 
& \heatG{89\%} & \heatG{64\%} & \heatG{68\%} & \heatDG{\textbf{100\%}} & \heatDG{90\%} & \heatG{78\%} & \heatG{62\%} & \heatDG{99\%} & \heatG{80\%} & \heatG{86\%} & \heatDG{\textbf{100\%}} & \heatDG{94\%} \\

Qwen3-8B
& & 
& \heatG{86\%} & \heatG{68\%} & \heatG{67\%} & \heatDG{\textbf{100\%}} & \heatDG{97\%} & \heatG{81\%} & \heatG{\textbf{65\%}} & \heatDG{98\%} & \heatG{83\%} & \heatDG{\textbf{93\%}} & \heatDG{\textbf{100\%}} & \heatDG{95\%} \\

\midrule
InternVL3.5-8B
& \multirow{4}{*}{560$\times$480}
& \multirow{4}{*}{p1}
& \heatDG{97\%} & \heatG{62\%} & \heatLG{57\%} & \heatDG{99\%} & \heatG{89\%} & \heatG{63\%} & \heatLG{55\%} & \heatDG{97\%} & \heatG{80\%} & \heatG{72\%} & \heatDG{99\%} & \heatDG{91\%} \\

Ovis2.5-9B
& & 
& \heatDG{92\%} & \heatG{65\%} & \heatG{67\%} & \heatDG{96\%} & \heatDG{97\%} & \heatG{82\%} & \heatLG{57\%} & \heatDG{98\%} & \heatG{80\%} & \heatG{88\%} & \heatDG{100\%} & \heatDG{93\%} \\

Qwen2.5-7B
& & 
& \heatG{72\%} & \heatLG{58\%} & \heatG{60\%} & \heatDG{98\%} & \heatDG{92\%} & \heatG{70\%} & \heatLG{57\%} & \heatDG{98\%} & \heatG{73\%} & \heatG{77\%} & \heatDG{99\%} & \heatG{85\%} \\

Qwen3-8B
& & 
& \heatG{78\%} & \heatG{60\%} & \heatLG{54\%} & \heatDG{100\%} & \heatDG{97\%} & \heatG{74\%} & \heatLG{57\%} & \heatDG{99\%} & \heatG{82\%} & \heatDG{90\%} & \heatDG{99\%} & \heatG{84\%} \\

\midrule
InternVL3.5-8B
& \multirow{4}{*}{224$\times$224}
& \multirow{4}{*}{p2}
& \heatG{88\%} & \heatG{69\%} & \heatG{61\%} & \heatDG{100\%} & \heatG{79\%} & \heatG{73\%} & \heatLG{57\%} & \heatDG{100\%} & \heatG{80\%} & \heatG{70\%} & \heatDG{100\%} & \heatDG{90\%} \\

Ovis2.5-9B
& & 
& \heatG{73\%} & \heatG{65\%} & \heatG{67\%} & \heatDG{100\%} & \heatDG{97\%} & \heatG{70\%} & \heatLG{57\%} & \heatDG{100\%} & \heatG{87\%} & \heatG{83\%} & \heatDG{100\%} & \heatDG{96\%} \\

Qwen2.5-7B
& & 
& \heatG{78\%} & \heatLG{57\%} & \heatLG{56\%} & \heatDG{94\%} & \heatG{79\%} & \heatG{63\%} & \heatLG{55\%} & \heatG{85\%} & \heatG{74\%} & \heatG{67\%} & \heatG{73\%} & \heatG{88\%} \\

Qwen3-8B
& & 
& \heatLG{55\%} & \heatG{62\%} & \heatLG{58\%} & \heatDG{100\%} & \heatG{88\%} & \heatO{50\%} & \heatG{60\%} & \heatDG{98\%} & \heatG{80\%} & \heatG{80\%} & \heatDG{100\%} & \heatDG{95\%} \\
\bottomrule
\end{tabular}%
}
\end{table*}

\subsection{Influence of Prompt Formulation}

We use a standardized prompt template to evaluate the vision-language model (VLM) in a constrained multiple-choice setting. 
The prompt consists of three components: (i) an image placeholder, (ii) a natural language question, and (iii) a set of answer options.

\begin{promptbox}{Perceptual Prompt (p1)}
{\small\ttfamily\raggedright
\prompttok{\textless image\textgreater} Do you see the \promptvar{\{object\}}? If yes, specify where: ``left'' or ``right''.\par
\medskip
A left\par
B right\par
\medskip
Answer with the option's letter from the given choices directly.
}
\end{promptbox}

\paragraph{Answer Format.}
The model is instructed to output only the corresponding option letter (e.g., ``A'' or ``B''), which ensures consistent and easily comparable predictions.

We also evaluate an alternative prompt that is closer to the VLA instruction format:

\begin{promptbox}{Action-Style Prompt (p2)}
{\small\ttfamily\raggedright
\prompttok{\textless image\textgreater} Put cube on \promptvar{\{object\}}\par
\medskip
A left\par
B right\par
\medskip
Answer with the option's letter from the given choices directly.
}
\end{promptbox}

\noindent As shown in Table~\ref{tab:appendix_vlm_resolution_prompt_ablation}, the action-style prompt slightly reduces VLM performance for some models and categories.
This effect is most visible for Qwen3-8B and Qwen2.5-7B, but the overall comparison remains qualitatively similar across prompt styles.
Nevertheless, VLM performance under the action-style prompt remains substantially above most VLA results in many knowledge-sensitive categories.

\subsection{Robustness to Texture Rendering}

We next evaluate whether the visual appearance of the simulated environment changes the main conclusions.
In the \emph{Raw Sim} setting, models are evaluated with the default simulator rendering, using the original simulated textures, materials, and backgrounds.
In the \emph{Visual Matching} setting \cite{simpler}, we follow the general visual-gap reduction strategy used in Simpler: simulated observations are made closer to real-world robot scenes by combining real-background compositing with foreground texture tuning for salient assets.
This includes matching or baking object textures from real images where possible and adjusting robot or scene textures that otherwise create a noticeable real-to-sim appearance gap.
Table~\ref{tab:appendix_environment_setup_ablations} shows that this change has only limited effect on the selected categories.
The same category-level pattern is preserved: \textsc{Color} and \textsc{Shape} remain easier than \textsc{Emotion} and \textsc{Attribute}, and Magma remains stronger than OpenVLA and $\pi_0$ on the tested semantic categories.

\begin{table}[t]
\centering
\scriptsize
\setlength{\tabcolsep}{2pt}
\renewcommand{\arraystretch}{0.86}

\caption{Environment setup ablations on representative categories.}
\label{tab:appendix_environment_setup_ablations}

\resizebox{\columnwidth}{!}{%
\begin{tabular}{lllcccc}
\toprule
\textbf{Abl.} & \textbf{Setting} & \textbf{Model}
& \textbf{Emo.}
& \textbf{Attr.}
& \textbf{Color}
& \textbf{Shape} \\
\midrule
\multirow{6}{*}{Texture}
& \multirow{3}{*}{Raw Sim}
& OpenVLA & \heatO{48\%} & \heatO{52\%} & \heatG{86\%} & \heatG{62\%} \\
& & $\pi_0$ & \heatO{50\%} & \heatO{48\%} & \heatG{86\%} & \heatO{51\%} \\
& & Magma & \heatG{71\%} & \heatG{63\%} & \heatDG{90\%} & \heatG{79\%} \\
\cmidrule(lr){2-7}
& \multirow{3}{*}{\begin{tabular}{@{}c@{}}Visual\\Matching\end{tabular}}
& OpenVLA & \heatO{48\%} & \heatO{51\%} & \heatG{89\%} & \heatG{64\%} \\
& & $\pi_0$ & \heatO{51\%} & \heatO{50\%} & \heatG{86\%} & \heatO{49\%} \\
& & Magma & \heatG{72\%} & \heatG{63\%} & \heatG{89\%} & \heatG{81\%} \\
\midrule
\multirow{9}{*}{Tile Size}
& \multirow{3}{*}{default}
& OpenVLA & \heatO{48\%} & \heatO{51\%} & \heatG{89\%} & \heatG{64\%} \\
& & $\pi_0$ & \heatO{51\%} & \heatO{50\%} & \heatG{86\%} & \heatO{49\%} \\
& & Magma & \heatG{72\%} & \heatG{63\%} & \heatG{89\%} & \heatG{81\%} \\
\cmidrule(lr){2-7}
& \multirow{3}{*}{+40\%}
& OpenVLA & \heatO{51\%} & \heatO{50\%} & \heatG{71\%} & \heatLG{56\%} \\
& & $\pi_0$ & \heatO{53\%} & \heatO{48\%} & \heatG{67\%} & \heatO{52\%} \\
& & Magma & \heatG{70\%} & \heatLG{59\%} & \heatDG{92\%} & \heatG{78\%} \\
\cmidrule(lr){2-7}
& \multirow{3}{*}{-20\%}
& OpenVLA & \heatO{52\%} & \heatO{49\%} & \heatG{86\%} & \heatG{65\%} \\
& & $\pi_0$ & \heatO{52\%} & \heatO{50\%} & \heatG{85\%} & \heatO{52\%} \\
& & Magma & \heatG{69\%} & \heatG{60\%} & \heatG{89\%} & \heatG{77\%} \\
\midrule
\multirow{9}{*}{Lighting}
& \multirow{3}{*}{default}
& OpenVLA & \heatO{48\%} & \heatO{51\%} & \heatG{89\%} & \heatG{64\%} \\
& & $\pi_0$ & \heatO{51\%} & \heatO{50\%} & \heatG{86\%} & \heatO{49\%} \\
& & Magma & \heatG{72\%} & \heatG{63\%} & \heatG{89\%} & \heatG{81\%} \\
\cmidrule(lr){2-7}
& \multirow{3}{*}{+30\%}
& OpenVLA & \heatO{51\%} & \heatO{48\%} & \heatG{86\%} & \heatLG{59\%} \\
& & $\pi_0$ & \heatO{47\%} & \heatO{50\%} & \heatG{78\%} & \heatO{50\%} \\
& & Magma & \heatG{71\%} & \heatG{60\%} & \heatDG{91\%} & \heatG{81\%} \\
\cmidrule(lr){2-7}
& \multirow{3}{*}{-30\%}
& OpenVLA & \heatO{49\%} & \heatO{47\%} & \heatDG{90\%} & \heatLG{54\%} \\
& & $\pi_0$ & \heatO{50\%} & \heatO{49\%} & \heatG{83\%} & \heatO{50\%} \\
& & Magma & \heatG{64\%} & \heatLG{58\%} & \heatG{89\%} & \heatG{80\%} \\
\bottomrule
\end{tabular}
}
\end{table}

\subsection{Effect of Answer-Tile Size}

We also test whether performance is sensitive to the physical size of the answer tiles.
This ablation is important because tile size may act as an out-of-distribution shift for VLA models.
As shown in Table~\ref{tab:appendix_environment_setup_ablations}, moderate changes preserve the main qualitative trends.
Increasing tile size can reduce performance for some models and categories, especially OpenVLA and $\pi_0$ on \textsc{Color}, but the broader pattern remains stable: simple perceptual categories are still easier than the more semantic categories, and the relative ranking among the tested models is largely unchanged.

\subsection{Effect of Lighting Intensity}

Finally, we vary lighting intensity to test whether results are driven by a narrow rendering condition.
Table~\ref{tab:appendix_environment_setup_ablations} shows that lighting perturbations can cause localized changes, especially under darker lighting and in harder categories.
However, the main conclusions are preserved: \textsc{Emotion} and \textsc{Attribute} remain difficult for OpenVLA and $\pi_0$, \textsc{Color} remains comparatively accessible, and Magma remains the strongest of the tested models on these categories.
These results suggest that the evaluation setup is not tied to a single lighting condition, while also confirming that perceptual rendering choices can affect some categories and should be controlled in future evaluations.

\section{Benchmark Sources and Data Construction}
\label{sec:appendix_benchmark_details}

Act2Answer is constructed from five source benchmarks: MLLM-CompBench~\cite{MLLM_bench}, IconQA~\cite{iconqa}, MMBench~\cite{mmbench}, OK-VQA~\cite{okvqa}, and VL-Think~\cite{blindvla}. 
We selected these benchmarks because together they provide established and complementary coverage of the knowledge domains targeted by Act2Answer. Our goal was to map these heterogeneous source benchmarks into a single evaluation setup: a short action-compatible instruction, two visual answer options, and a common embodied binary action-selection protocol.

The degree of adaptation therefore differs by source benchmark and is driven by source-format mismatch rather than arbitrary redesign. In practice, we prioritized source tasks that could be converted into embodied binary choice without substantially changing what the original task was meant to test. We further restricted the final pool to examples whose relevant visual evidence remains perceivable under the lower effective resolution typical of current VLA systems and whose answer can be expressed through a short action-conditioned instruction.

\begin{figure*}[!thb]
    \centering
    \includegraphics[width=\textwidth]{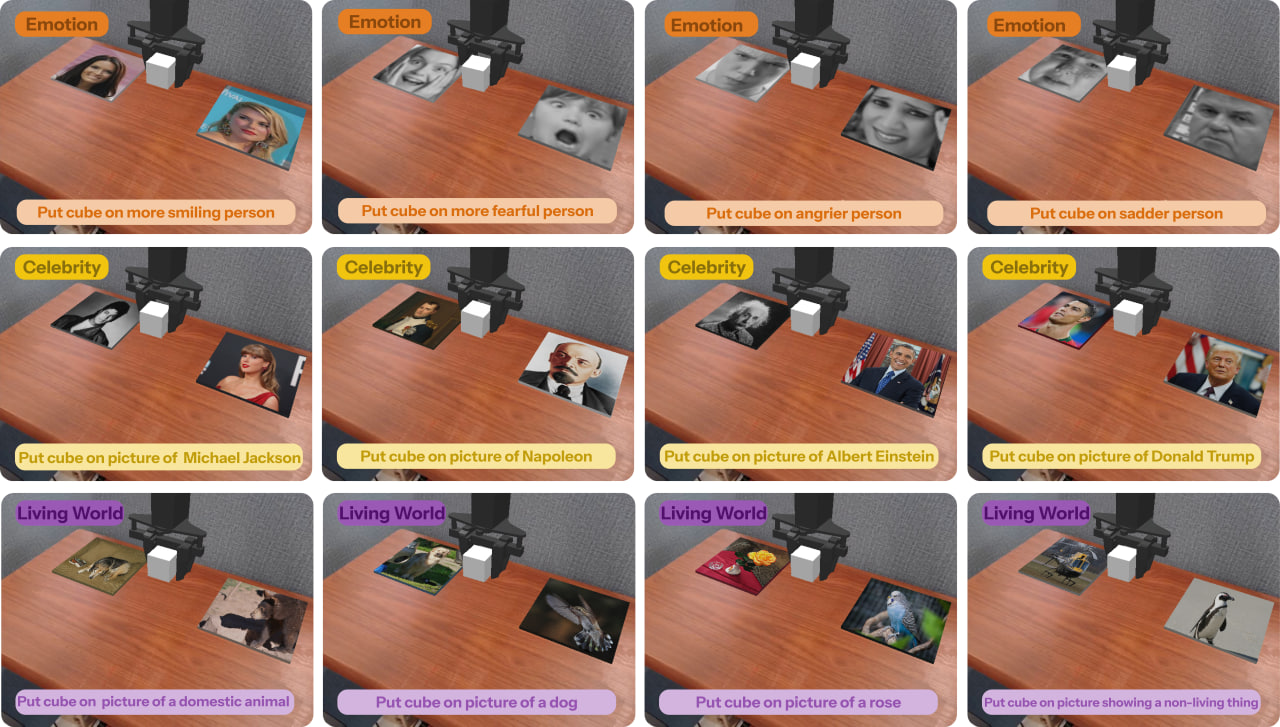}
    \caption{Additional \textsc{Act2Answer} environment examples from the \textsc{Emotion}, \textsc{Celebrity}, and \textsc{Living World} categories.}
    \label{fig:appendix_examples_social_bio}
\end{figure*}

\begin{figure*}[!thb]
    \centering
    \includegraphics[width=\textwidth]{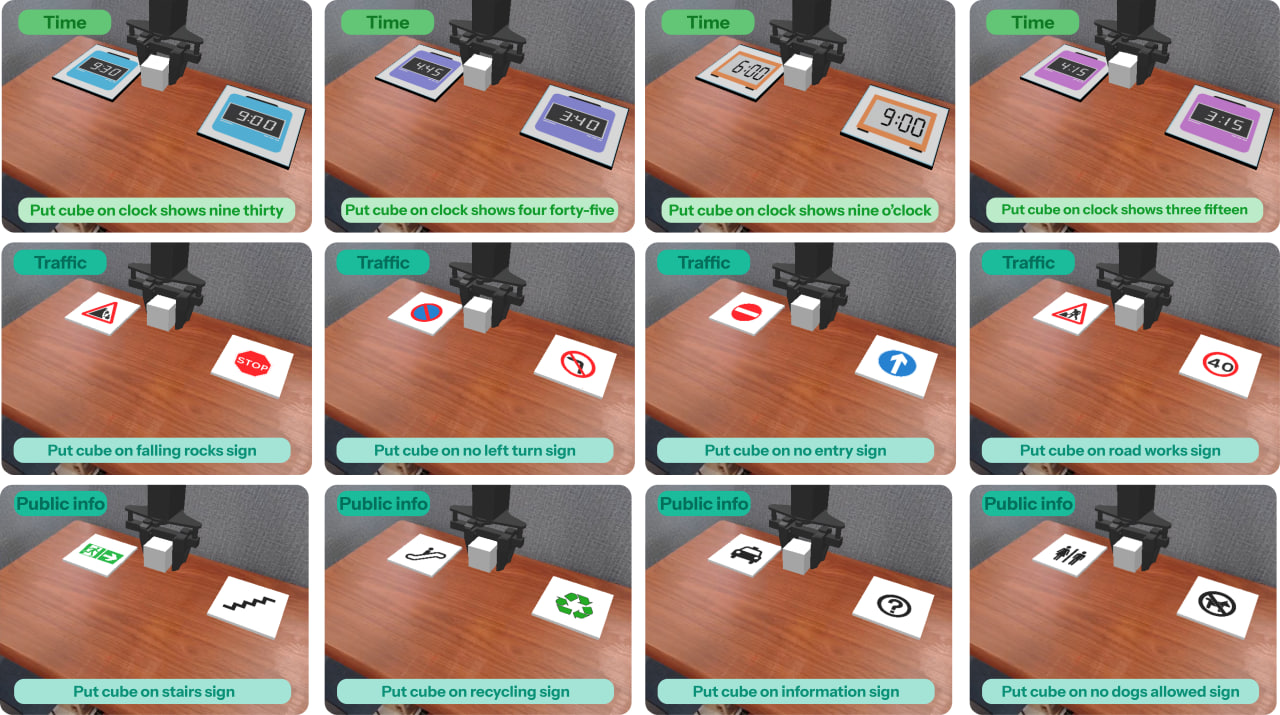}
    \caption{Additional \textsc{Act2Answer} environment examples from the \textsc{Time}, \textsc{Traffic}, and \textsc{Public info} categories.}
    \label{fig:appendix_examples_time_norm}
\end{figure*}

\begin{figure*}[!hbt]
    \centering
    \includegraphics[width=\linewidth]{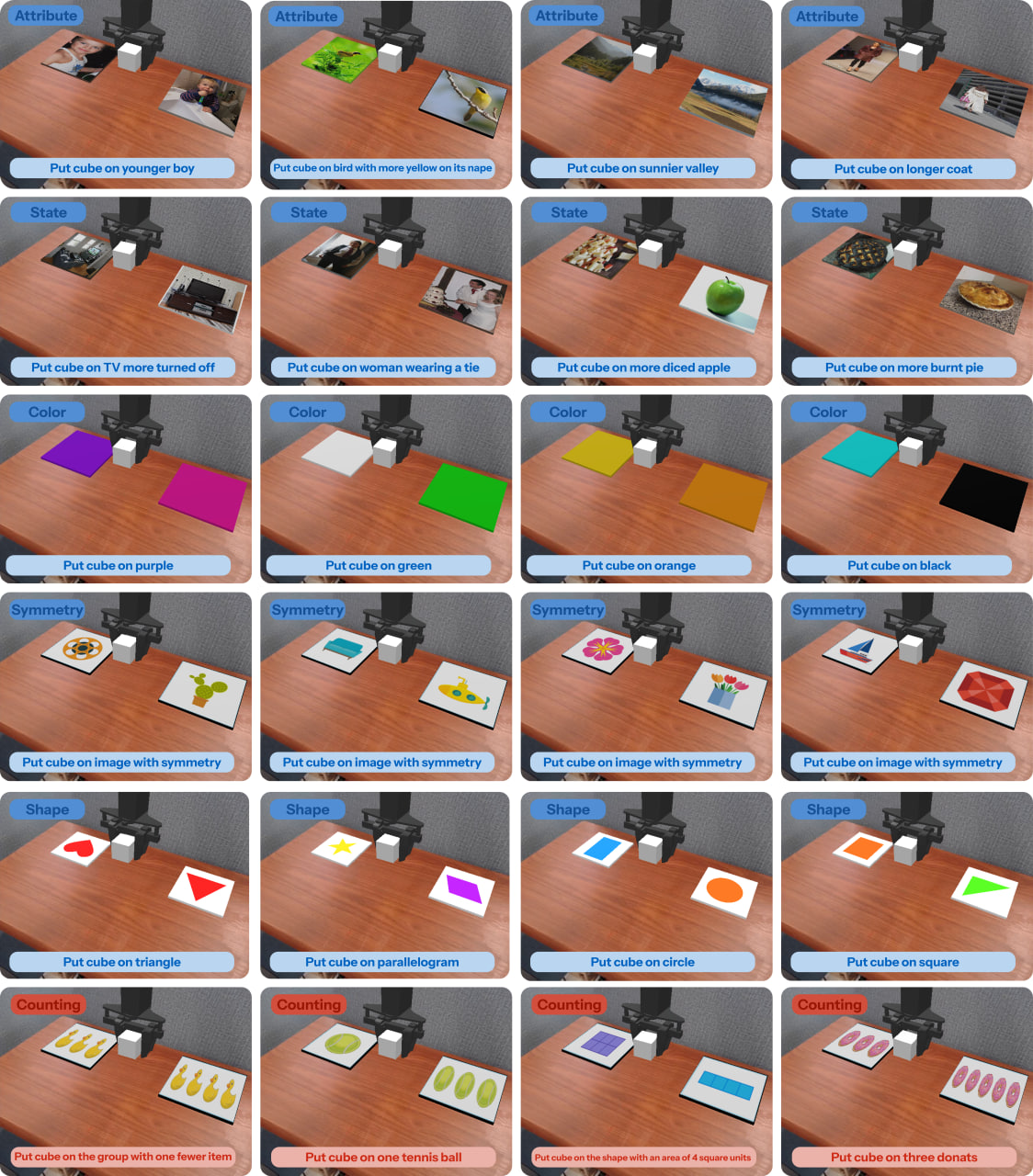}
    \caption{Additional \textsc{Act2Answer} environment examples from the \textsc{Attribute}, \textsc{State}, \textsc{Color}, \textsc{Symmetry}, \textsc{Shape}, and \textsc{Counting} categories.}
    \label{fig:appendix_examples_phys_quant}
\end{figure*}

Representative examples of the resulting embodied tasks are shown in Figures~\ref{fig:appendix_examples_social_bio}, \ref{fig:appendix_examples_time_norm}, and \ref{fig:appendix_examples_phys_quant}. For readability, we group them into three panels: (i) social, biological, and culturally grounded categories, (ii) temporal and public-convention categories, and (iii) physical and quantitative categories.

\FloatBarrier

\begin{table*}[h!]
\centering
\small
\begin{tabular}{llllrr}
\toprule
\textbf{Source benchmark} & \textbf{Domain} & \textbf{Native format} & \textbf{Transfer type} & \textbf{Initial pool} & \textbf{Final eval} \\
\addlinespace
\midrule
\addlinespace
\multirow{3}{*}{MLLM-CompBench}& Emotion & \multirow{3}{*}{image pair + comparative question} & \multirow{3}{*}{near} & 5254 & 300 \\ & Attribute & & & 5386 & 300 \\ & State & & & 1066 & 300 \\
\addlinespace
\midrule
\addlinespace
\multirow{4}{*}{IconQA} & Time & \multirow{4}{*}{multi-image choice} & \multirow{4}{*}{moderate} & 508 & 300 \\ & Shape & & & 12798 & 300 \\ & Symmetry  & & & 362 & 300 \\ & Counting  & & & 5001 & 300 \\
\addlinespace
\midrule
\addlinespace
MMBench & Celebrity & multiple choice & moderate & 316 & 140 \\
\midrule
OK-VQA & Living World & open-ended VQA & heavy & 2336 & 300 \\
\addlinespace
\midrule
\addlinespace
\multirow{3}{*}{VL-Think} & Public Info & \multirow{3}{*}{embodied board selection} & \multirow{3}{*}{near} & 14 concepts & 300 \\ & Traffic & & & 24 concepts & 300 \\ & Color & & & 11 concepts & 300 \\
\bottomrule
\end{tabular}
\caption{Per-category curation statistics for the \textsc{Act2Answer} suite. `Final eval' reports the total number of evaluation episodes after including both the original and swapped left/right configurations for each selected item. For VL-Think, the `Initial pool' column reports the number of source concepts rather than the number of candidate image examples.}
\label{tab:appendix_per_category_stats}
\end{table*}

\subsection{Near Format-Preserving Adaptations}

\noindent\textit{MLLM-CompBench.}
MLLM-CompBench was the cleanest source benchmark for Act2Answer. Its native format already consists of two candidate images and a comparative question, making it closely aligned with our final embodied answer-selection setup. We therefore used it as the primary source for \textsc{Emotion}, \textsc{Attribute}, and \textsc{State}. Adaptation in this case was near format-preserving: we retained the original image pair and converted the question into a short action-compatible instruction. Because almost no additional restructuring was required, these categories provide one of the most direct comparisons between VLM and VLA behavior in our study.

\noindent\textit{VL-Think.}
VL-Think was also a natural fit for our protocol because its task structure is already close to an embodied semantic selection problem. We used it for \textsc{Public Info}, \textsc{Traffic}, and \textsc{Color}, where the relevant concepts are encoded in compact visual symbols or simple public conventions. Since the source benchmark already operates in a board-selection style setting, adaptation remained near format-preserving and mainly consisted of unifying the instruction style and episode format with the rest of the Act2Answer suite. Because VL-Think is specified in terms of small closed concept vocabularies rather than large candidate image pools, the corresponding `Initial pool' entries in Table~\ref{tab:appendix_per_category_stats} report the number of source concepts rather than the number of raw image examples.

\subsection{Moderately Adapted Benchmarks}

\noindent\textit{IconQA.}
IconQA was used for \textsc{Time}, \textsc{Shape}, \textsc{Symmetry}, and \textsc{Counting}. It required moderate adaptation because its native format is multi-option visual choice rather than binary embodied selection. We therefore converted each item into binary choice by pairing the correct answer with one distractor at a time while preserving the original semantic target. Because IconQA uses schematic icon diagrams rather than cluttered natural images, it remains relatively robust under the lower visual resolution typical of current VLA systems. We retained only those subsets that fit the Act2Answer protocol most naturally, including geometry-based subsets for \textsc{Shape} and \textsc{Symmetry}, and a visually grounded clock-style subset for \textsc{Time}.

\noindent\textit{MMBench.}
We used MMBench for the \textsc{Celebrity} category. This subset required moderate adaptation. We kept the same identity-recognition task, but re-curated the set of public figures to focus on more broadly recognizable identities. The aim was to better target shared world knowledge without changing the underlying recognition problem. The resulting subset is therefore better understood as a curated derivative of MMBench rather than a verbatim extraction.

\subsection{Open-Ended to Binary Adaptation}

\noindent\textit{OK-VQA.}
OK-VQA was used for the \textsc{Living World} category, covering animal identity, flora identity, and living-versus-nonliving distinctions. This benchmark required the most benchmark-specific adaptation because its native format is single-image open-ended VQA, whereas Act2Answer evaluates knowledge through binary embodied selection. We first retained only examples with short, visually grounded answers and stable annotator agreement. We then selected biologically relevant cases through question-pattern and answer-level filtering. To convert these items into the common format, each source image was paired with a second image from the same category that did not contain the correct answer. This allowed the same biological knowledge to be tested through embodied binary choice. Ambiguous cases were removed, and the final embodied instructions were edited accordingly. The resulting OK-VQA-derived subset should be understood as a benchmark-specific adaptation from open-ended VQA to embodied binary evaluation.

\subsection{Instruction Rewriting and Binary Conversion}
\label{sec:appendix_instruction_rewriting}

Our goal was to preserve the original knowledge requirement while changing the response modality from text to action. We therefore used an LLM as a first-pass rewriting tool to convert source benchmark questions into short imperative instructions for VLA models, followed by human review and manual editing. This rewriting was limited to template normalization rather than semantic reformulation. For example, comparative prompts from MLLM-CompBench were converted into instructions of the form \textit{``Put the cube on \ldots''} while preserving the same comparison target and candidate images.

Prompts followed a simple instruction style, e.g., \textit{``Put the cube on the more smiling person''} (\textsc{Emotion}), \textit{``Put the cube on dryer grass''} (\textsc{State}), and \textit{``Put the cube on the picture of sheep''} (\textsc{Living World}). Throughout the suite, instructions were kept short, visually grounded, and close to the original semantic target.

\section{Discussion}

\paragraph{How to prevent the erosion of world knowledge?}
Finally, it is not yet clear how to systematically \emph{prevent the erosion of world knowledge during VLA training}. 
Developing effective training procedures that retain the backbone’s world knowledge while acquiring strong control policies remains an open challenge.
Promising directions include multi-task and continual-learning schemes that interleave action prediction with general question answering~\cite{zitkovich2023rt, amin2025pi, ivanova-etal-2025-ambik}, regularization or distillation techniques that explicitly preserve VLM representations, and architectural decoupling of knowledge and control components. Another complementary direction is inference-time intervention on interpretable internal features, such as SAE-based steering toward clarification behavior in ambiguous embodied-instruction settings~\citep{petrova2026steering}.

To make this discussion more concrete, we also ran preliminary mitigation experiments on a small subset of representative \textsc{Act2Answer} categories, shown in Table~\ref{tab:appendix_mitigation_ablation}.
These experiments are not intended as full mitigation methods, but as simple probes of two plausible directions.
First, we test language-rephrasing augmentation during downstream fine-tuning, where instructions are varied through action-verb substitutions, spatial-expression variants, sentence-structure changes, descriptive noun replacements, and robot-directed commands.
Second, we test a latent-distillation baseline that adds a representation-preservation loss between mid-layer VLA hidden states and final patch embeddings from a frozen vision foundation teacher.
Both variants are obtained by fine-tuning $\pi_0$ on BridgeDataV2 pick-and-place data and do not use \textsc{Act2Answer} evaluation examples for training.

The results suggest that these lightweight interventions are useful but limited.
Language rephrasing slightly improves \textsc{Color} and \textsc{Shape}, but does not improve the more semantic \textsc{Emotion} and \textsc{Attribute} categories.
Latent distillation shows a similar pattern: it improves \textsc{Shape} and preserves strong \textsc{Color} performance, but still leaves \textsc{Emotion} and \textsc{Attribute} near chance.
This pattern is consistent with the main findings of the paper: simple perceptual distinctions are easier to preserve or recover, whereas richer semantic distinctions require stronger mechanisms than shallow instruction diversity or a single representation-matching objective.
At the same time, the distillation result points to a promising direction for future work, especially representation-preservation losses during VLA pretraining, better choices of teacher representations and distillation data, and comparisons between rehearsal, regularization, and parameter-efficient adaptation.

However, large-scale multi-task or continual training that keeps many skills active can be prohibitively compute-intensive for real-world systems, especially when combined with high-resolution perception and long-horizon control.
This makes \textit{it essential to explore more efficient paradigms and lightweight mechanisms to prevent forgetting} - such as parameter-efficient fine-tuning, targeted regularization, selective rehearsal, or sparsely updated knowledge modules - so that VLAs can become more capable without repeatedly sacrificing the commonsense they started with.

\begin{table}[t]
\centering
\footnotesize
\setlength{\tabcolsep}{3pt}
\renewcommand{\arraystretch}{0.95}

\caption{Preliminary mitigation ablations on a subset of representative categories.}
\label{tab:appendix_mitigation_ablation}

\begin{tabular}{lcccc}
\toprule
\textbf{Model}
& \textbf{Emotion}
& \textbf{Attribute}
& \textbf{Color}
& \textbf{Shape} \\
\midrule

$\pi_0$
& \heatO{51\%} & \heatO{50\%} & \heatG{86\%} & \heatO{49\%} \\
$\pi_0$ + paraphrases
& \heatO{47\%} & \heatO{48\%} & \heatG{89\%} & \heatO{53\%} \\
$\pi_0$ + distill
& \heatO{49\%} & \heatO{50\%} & \heatG{88\%} & \heatLG{56\%} \\

\bottomrule
\end{tabular}
\end{table}

\begin{table*}[t]
\centering
\footnotesize
\setlength{\tabcolsep}{3.5pt}
\renewcommand{\arraystretch}{0.95}

\caption{Left/right split results for selected VLA models with heatmap formatting.}
\label{tab:act2answer_left_right_split_heatmap}

\resizebox{\textwidth}{!}{%
\begin{tabular}{l cc!{\vrule width 0.6pt} cc!{\vrule width 0.6pt} cc!{\vrule width 0.6pt} cc!{\vrule width 0.6pt} cc!{\vrule width 0.6pt} cc!{\vrule width 0.6pt} cc!{\vrule width 0.6pt} cc!{\vrule width 0.6pt} cc!{\vrule width 0.6pt} cc!{\vrule width 0.6pt} cc!{\vrule width 0.6pt} cc}
\toprule
\textbf{Model}
& \multicolumn{2}{c!{\vrule width 0.6pt}}{\textbf{Social}}
& \multicolumn{10}{c!{\vrule width 0.6pt}}{\textbf{Physical}}
& \multicolumn{2}{c!{\vrule width 0.6pt}}{\textbf{Quantitative}}
& \multicolumn{2}{c!{\vrule width 0.6pt}}{\textbf{Temporal}}
& \multicolumn{4}{c!{\vrule width 0.6pt}}{\textbf{Normative}}
& \multicolumn{2}{c!{\vrule width 0.6pt}}{\textbf{Cultural}}
& \multicolumn{2}{c}{\textbf{Biological}} \\
\cmidrule(lr){2-3}
\cmidrule(lr){4-13}
\cmidrule(lr){14-15}
\cmidrule(lr){16-17}
\cmidrule(lr){18-21}
\cmidrule(lr){22-23}
\cmidrule(lr){24-25}
& \multicolumn{2}{c!{\vrule width 0.6pt}}{\textbf{Emotion}}
& \multicolumn{2}{c!{\vrule width 0.6pt}}{\textbf{Attribute}}
& \multicolumn{2}{c!{\vrule width 0.6pt}}{\textbf{State}}
& \multicolumn{2}{c!{\vrule width 0.6pt}}{\textbf{Color}}
& \multicolumn{2}{c!{\vrule width 0.6pt}}{\textbf{Shape}}
& \multicolumn{2}{c!{\vrule width 0.6pt}}{\textbf{Symmetry}}
& \multicolumn{2}{c!{\vrule width 0.6pt}}{\textbf{Counting}}
& \multicolumn{2}{c!{\vrule width 0.6pt}}{\textbf{Time}}
& \multicolumn{2}{c!{\vrule width 0.6pt}}{\textbf{Public Info}}
& \multicolumn{2}{c!{\vrule width 0.6pt}}{\textbf{Traffic}}
& \multicolumn{2}{c!{\vrule width 0.6pt}}{\textbf{Celebrity}}
& \multicolumn{2}{c}{\textbf{Living World}} \\
\cmidrule(lr){2-3}
\cmidrule(lr){4-5}
\cmidrule(lr){6-7}
\cmidrule(lr){8-9}
\cmidrule(lr){10-11}
\cmidrule(lr){12-13}
\cmidrule(lr){14-15}
\cmidrule(lr){16-17}
\cmidrule(lr){18-19}
\cmidrule(lr){20-21}
\cmidrule(lr){22-23}
\cmidrule(lr){24-25}
& \textbf{L} & \textbf{R}
& \textbf{L} & \textbf{R}
& \textbf{L} & \textbf{R}
& \textbf{L} & \textbf{R}
& \textbf{L} & \textbf{R}
& \textbf{L} & \textbf{R}
& \textbf{L} & \textbf{R}
& \textbf{L} & \textbf{R}
& \textbf{L} & \textbf{R}
& \textbf{L} & \textbf{R}
& \textbf{L} & \textbf{R}
& \textbf{L} & \textbf{R} \\
\midrule

SpatialVLA
& \heatR{36\%} & \heatLG{58\%}
& \heatLG{55\%} & \heatR{42\%}
& \heatO{52\%} & \heatO{48\%}
& \heatG{80\%} & \heatDG{95\%}
& \heatG{82\%} & \heatG{85\%}
& \heatDG{90\%} & \heatR{0\%}
& \heatO{50\%} & \heatLG{55\%}
& \heatO{47\%} & \heatO{48\%}
& \heatG{68\%} & \heatR{35\%}
& \heatG{61\%} & \heatO{53\%}
& \heatG{66\%} & \heatR{44\%}
& \heatG{67\%} & \heatR{32\%} \\

Magma
& \heatG{76\%} & \heatG{68\%}
& \heatG{65\%} & \heatG{61\%}
& \heatG{62\%} & \heatLG{56\%}
& \heatDG{91\%} & \heatG{87\%}
& \heatG{81\%} & \heatG{86\%}
& \heatO{46\%} & \heatR{28\%}
& \heatO{52\%} & \heatO{52\%}
& \heatG{77\%} & \heatG{78\%}
& \heatDG{90\%} & \heatG{86\%}
& \heatG{89\%} & \heatG{71\%}
& \heatDG{94\%} & \heatDG{94\%}
& \heatG{81\%} & \heatG{73\%} \\

Xiaomi-Robotics-R0
& \heatG{66\%} & \heatG{60\%}
& \heatO{46\%} & \heatLG{58\%}
& \heatLG{54\%} & \heatO{46\%}
& \heatG{86\%} & \heatDG{95\%}
& \heatG{84\%} & \heatG{79\%}
& \heatO{46\%} & \heatG{70\%}
& \heatR{45\%} & \heatO{50\%}
& \heatG{65\%} & \heatR{39\%}
& \heatG{69\%} & \heatLG{59\%}
& \heatG{63\%} & \heatO{51\%}
& \heatG{70\%} & \heatG{65\%}
& \heatG{62\%} & \heatO{51\%} \\

InternVLA-M1
& \heatO{52\%} & \heatLG{54\%}
& \heatO{50\%} & \heatO{49\%}
& \heatLG{59\%} & \heatO{47\%}
& \heatDG{93\%} & \heatG{88\%}
& \heatG{80\%} & \heatO{52\%}
& \heatR{13\%} & \heatG{74\%}
& \heatO{50\%} & \heatO{47\%}
& \heatR{40\%} & \heatLG{58\%}
& \heatG{83\%} & \heatR{26\%}
& \heatG{87\%} & \heatR{20\%}
& \heatLG{54\%} & \heatO{50\%}
& \heatLG{54\%} & \heatG{63\%} \\

\bottomrule
\end{tabular}%
}
\end{table*}

\section{Details of Score Averaging}
\label{appendix:score_averaging}

For LIBERO, all values for the compared VLA models were taken directly from the original papers. For Act2Answer, we used the results from \autoref{tab:act2answer_emotion_attribute_state}. For each model, we first averaged accuracy across all environments, and then normalized this mean relative to random guessing (50\%) using a linear rescaling:
\begin{equation}
\text{Normalized Score} = \frac{\text{Mean Accuracy} - 50}{50}\times 100.
\end{equation}

\section{Chance Margin $\Delta$ and Significance Thresholds}
\label{appendix:delta}

The three interpretation regimes of the Soft Success Rate (Section~\ref{subsec:soft_success_rate}) are defined relative to a chance margin $\Delta$ around the random-guessing baseline of $0.5$. This appendix defines $\Delta$ and explains how its value is chosen.

Each \textsc{Act2Answer} item is a binary forced choice, so under a knowledge-free policy the per-episode outcome is a Bernoulli trial with success probability $p_0 = 0.5$. For a category evaluated over $N$ episodes, the empirical $\mathrm{SR}$ is a sample proportion whose standard error at chance is $\sqrt{p_0(1-p_0)/N} = \sqrt{0.25/N}$. We set $\Delta$ to the corresponding two-sided Wald confidence half-width,
\begin{equation}
\Delta = z_{1-\alpha/2}\,\sqrt{\frac{p_0(1-p_0)}{N}} = z_{1-\alpha/2}\,\sqrt{\frac{0.25}{N}},
\end{equation}
so that the band $|\mathrm{SR}-0.5| \le \Delta$ contains exactly those scores that are not significantly different from chance at level $\alpha$. We use $\alpha = 0.05$ ($z_{1-\alpha/2} = 1.96$) throughout. A category result is then counted as \emph{evidence of usable knowledge} when $\mathrm{SR} > 0.5 + \Delta$, as \emph{instruction or perceptual failure} when $\mathrm{SR} < 0.5 - \Delta$, and as \emph{no reliable usable knowledge} otherwise.

Because $\Delta$ depends on the number of evaluated episodes, it is computed per category. Most categories are evaluated over $N = 300$ episodes (150 items in both the original and swapped left/right configurations), which gives $\Delta \approx 0.057$ (about $5.7$ percentage points). The smaller \textsc{Celebrity} set uses $N = 140$ episodes, giving a wider band of $\Delta \approx 0.083$.

The two swapped views of a single item are not fully independent, so treating each episode as an independent trial is mildly anticonservative. Using the number of unique items as the effective sample size (e.g., $N = 150$ rather than $300$) yields a slightly wider, more conservative margin ($\Delta \approx 0.08$). Our qualitative conclusions are unchanged under either choice, since the gaps we report between primitive and richer semantic categories are substantially larger than $\Delta$.

\section{Left/Right Swapped-Layout Analysis}
\label{appendix:left_right_swapped_layout}

Act2Answer uses a binary tabletop layout, so a model with a fixed spatial preference could obtain misleading scores if each question were evaluated in only one candidate arrangement.
To control for this, each item is evaluated in two spatial configurations: the original left/right placement of the candidate answer images and a swapped configuration in which the two candidates exchange positions.
The main Act2Answer score averages over these two configurations, so systematic side preferences have less influence on the final category-level result.

Table~\ref{tab:act2answer_left_right_split_heatmap} reports the left/right breakdown for selected VLA models.
The split is useful because it makes positional effects visible rather than hiding them inside a single averaged score.
When a model does not have a reliable knowledge-conditioned signal for a category, performance can vary substantially across the two layouts, indicating either a side preference, a layout-specific failure, or near-random action selection.
For example, several models show large left/right asymmetries on harder categories such as \textsc{Symmetry}, \textsc{Public Info}, \textsc{Traffic}, and \textsc{Living World}.
In contrast, categories where performance remains high in both configurations, such as \textsc{Color} for most evaluated models, provide stronger evidence that the model is responding to the intended visual-semantic content rather than simply exploiting position.

This analysis supports the role of swapped-layout evaluation as a built-in robustness control in Act2Answer.
It does not by itself eliminate all spatial or perceptual confounds, since layout can still interact with model-specific perception and control behavior.
However, it reduces the effect of fixed side preferences on the reported score and provides a diagnostic view of when apparent success or failure may be driven by position rather than by knowledge-sensitive action selection.

\end{document}